\pdfoutput=1

\documentclass[11pt]{article}

\usepackage[]{acl}

\usepackage{times}
\usepackage{latexsym}

\usepackage{graphicx}
\usepackage{subfigure}
\usepackage{amsmath}
\usepackage{multirow}
\usepackage{caption}

\usepackage[T1]{fontenc}

\usepackage[utf8]{inputenc}

\usepackage{microtype}

\usepackage{hyperref}

\title{Enhance Incomplete Utterance Restoration by Joint Learning \\Token Extraction and Text Generation}

\author{Shumpei Inoue$^1$, Tsungwei Liu$^1$, Nguyen Hong Son$^1$, Minh-Tien Nguyen$^{1,2,}$\thanks{$^*$Corresponding Author.} \\
        $^1$Cinnamon AI, 10th floor, Geleximco building, 36 Hoang Cau, Dong Da, Hanoi, Vietnam. \\
        \texttt{\{sinoue, tsungwei.liu, levi, ryan.nguyen\}@cinnamon.is} \\
        $^2$Hung Yen University of Technology and Education, Hung Yen, Vietnam. \\
        \texttt{tiennm@utehy.edu.vn}}

\begin{document}
\maketitle
\begin{abstract}
This paper introduces a model for incomplete utterance restoration (IUR) called JET (\textbf{J}oint learning token \textbf{E}xtraction and \textbf{T}ext generation). Different from prior studies that only work on extraction or abstraction datasets, we design a simple but effective model, working for both scenarios of IUR. Our design simulates the nature of IUR, where omitted tokens from the context contribute to restoration. From this, we construct a Picker that identifies the omitted tokens. To support the picker, we design two label creation methods (soft and hard labels), which can work in cases of no annotation data for the omitted tokens. The restoration is done by using a Generator with the help of the Picker on joint learning. Promising results on four benchmark datasets in extraction and abstraction scenarios show that our model is better than the pretrained T5 and non-generative language model methods in both rich and limited training data settings.\footnote{The code is available at \url{https://github.com/shumpei19/JET}}

\end{abstract}

\section{Introduction}\label{sec:intro}



\begin{figure*}[!h]
	\centering
	\includegraphics[width=1.0\textwidth]{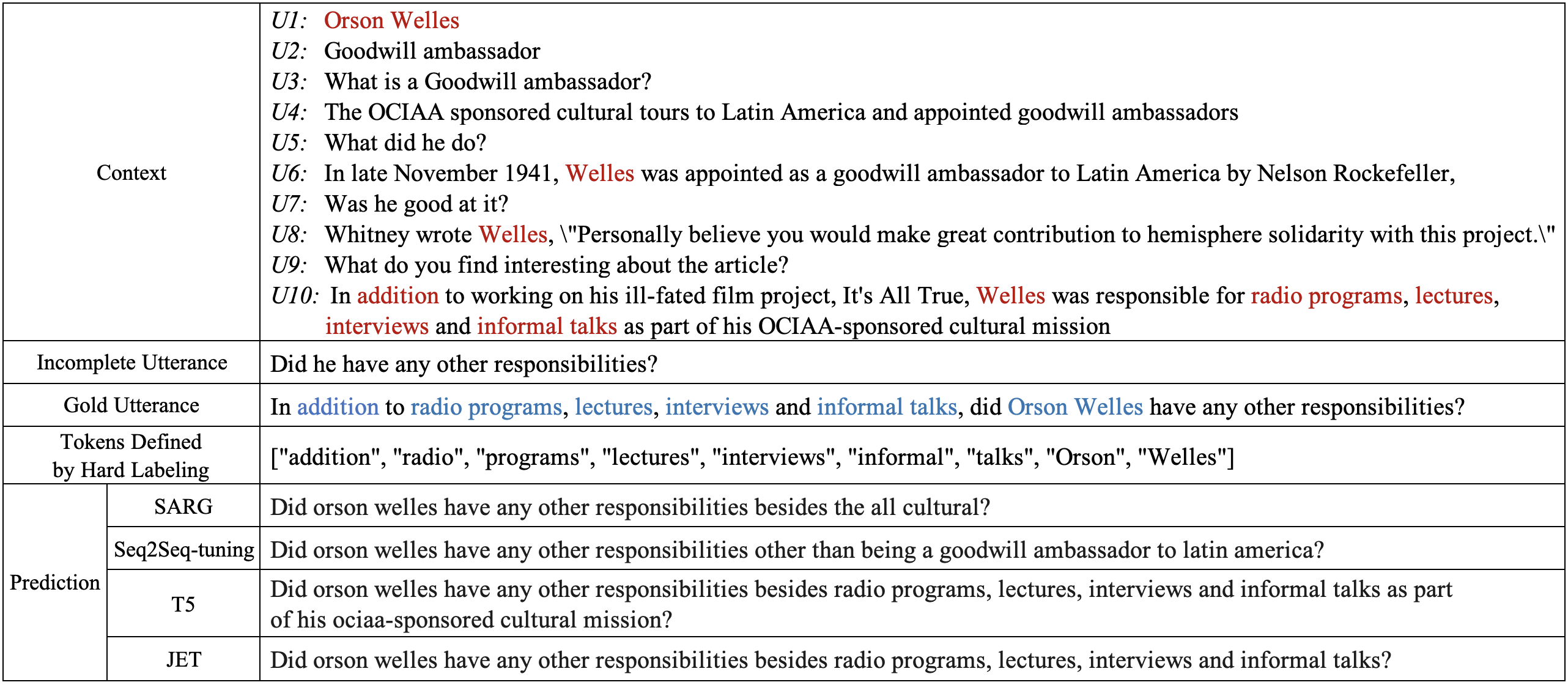}
	\caption{The sample data from \textit{CANARD}. IUR models rewrite the incomplete utterance to be as similar as possible to the reference. The blue tokens are omitted tokens (excluding stop words) in the incomplete utterance. The red tokens are defined by our hard labeling approach as an important token.}\label{fig:example}
\end{figure*}

Understanding conversational interactions through NLP has become important with increasing connectivity and range of capabilities. The applications using natural conversations cover a wide range of solutions including dialogue systems, information extraction, and summarization. For example, \citet{Adiwardana-human-chatbot-20, Su-Movie-chat-EMNLP-20} aimed to build the dialogue system where an intelligent virtual agent answers human conversations and makes suggestions in an open/closed domain. \citet{Bak-king-decision-EMNLP-18,Karan-Topic-bias-SIGDIAL-21} attempted to detect decision-related utterances from multi-party meeting recordings, while \citet{tarnpradab2017toward} applied extractive summarization for online forum discussions. These features allow users to to quickly catch up with the current situation, decisions and next-action without having to follow a lengthy or comprehensive dialogue. However, utterances, the components of a conversation, are generally not self-contained and are difficult to understand by their own. This comes from the nature of multi-turn dialogue where each utterance contains co-references, rephrases, and ellipses (Figure \ref{fig:example}). \citeauthor{Su-TPtr-ACL-19} \citeyear{Su-TPtr-ACL-19} also showed that co-references and ellipses occur in over 70\% of utterances in conversations. This is a ubiquitous problem in conversational AI, making the challenge for building practical systems with conversations.



 
Incomplete Utterance Restoration (IUR) \cite{Pan-Restoration-EMNLP-19} is one solution to restore semantically underspecified  utterances (i.e., incomplete utterances) in conversations. Figure \ref{fig:example} shows an example of IUR, in which the model rewrites the incomplete utterance to the reference. IUR is a challenging task due to two reasons. Firstly, the gold utterance (the reference) overlaps a lot of tokens with the pre-restored, incomplete utterance, while it overlaps only a few tokens with utterances in the context. We observed that for CANARD \cite{Elgohary-CANARD-EMNLP-19}, 85\% of tokens in incomplete utterances were directly cited for rewriting, while only 17\% of tokens in context was cited for rewriting. Secondly, it is important to detect omitted tokens in incomplete utterances and to include them in the restoration process. In actual cases of IUR, no matter how fluent and grammatically correct the machine's generation is, it is useless as long as important tokens are left out.




Recent studies used several methods for IUR. It includes the extraction of omitted tokens for restoration (PAC) \cite{Pan-Restoration-EMNLP-19}, two-stage learning \cite{Song-Multi-task-WC}, seq2seq fine-tuning \cite{bao2021s2s}, semantic segmentation (RUN-BERT) \cite{Liu-RUN-BERT-EMNLP-20}, or the tagger to detect which tokens in incomplete utterances should be kept, deleted or changed for restoration (SARG) \cite{Huang-SARG-AAAI-21}. However, we argue that these methods can only work on neither extractive nor abstractive IUR datasets. For example, SARG and seq2seq achieve promising results on \textit{Restoration 200k} \cite{Pan-Restoration-EMNLP-19} where omitted tokens can be directly extracted from the context (extraction). But they are not the best on CANARD \cite{Elgohary-CANARD-EMNLP-19}, which requires more abstraction for restoration. In Figure \ref{fig:example},\footnote{The performance of RUN-BERT is limited on CANARD.} we can observe that the output of SARG and seq2seq are worse than that of our JET. Text editing strategy by SARG is limited in its ability to generate abstractive rewriting while seq2seq has the problem in picking omitted tokens. As the result, the generality of these methods is still an open question.





We introduce a simple but effective model to deal with the generality of IUR methods named \textbf{JET} (\textbf{J}oint learning token \textbf{E}xtraction and \textbf{T}ext generation). The model is designed to work widely from extractive to abstractive scenarios. To do that, we first address the problem of identifying omitted tokens from the dialogue context by introducing a picker. The picker uses a new matching method for dealing with various forms of tokens (Figure \ref{fig:example}) in the extraction style. We next consider the abstraction aspect of restoration by offering a generator. The generator utilizes the power of the pre-trained T5 model to rewrite incomplete utterances. The picker and generator share the T5's encoder and are jointly trained in a unified model for IUR.
This paper makes three main contributions:
\begin{itemize}
    \item We propose JET, a simple but effective model based on T5 for utterance restoration in multi-turn conversations. Our model jointly optimizes two tasks: picking important tokens (the picker) and generating re-written utterances (the generator). To our best knowledge, we are the first to utilize T5 for the IUR task.
    
    
    
    \item We design a method for identifying important tokens for training the picker. The method facilitates IUR models in actual cases, in which there are no (a few) existing gold labels.
    
    \item We demonstrate the validity of the model by comparing it to strong baselines from multiple perspectives such as limited data setting (Section \ref{sec:train-lim}), human evaluation (Section \ref{sec:human-eval}) and output observation (Section \ref{sec:output-obs}).
    
\end{itemize}


\begin{figure*}[!h]
	\centering
	\includegraphics[width=1.0\textwidth]{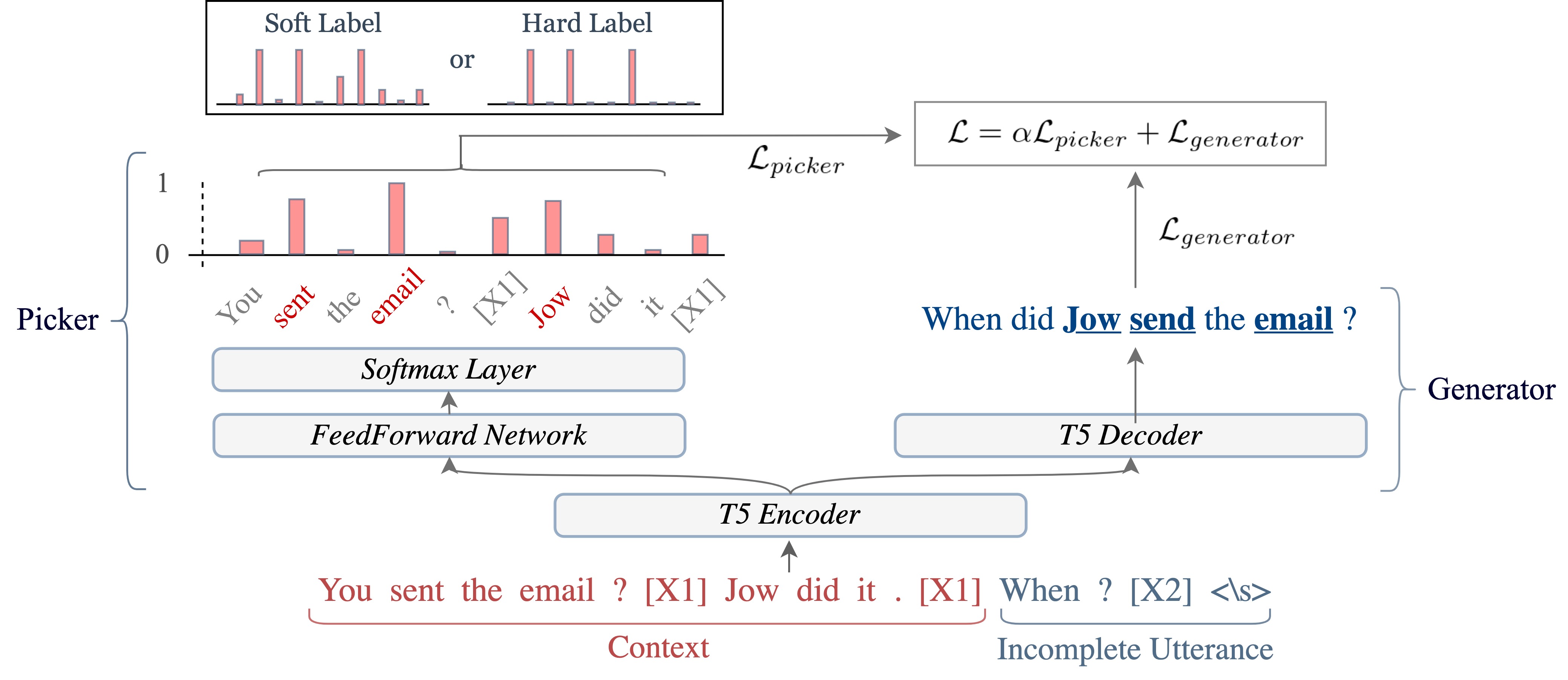}
	\caption{The proposed model (JET) for utterance restoration. The left is the picker and the right is the generator. The model jointly optimizes two tasks for doing restoration. The input format for our model is described in \ref{subsec:input} }\label{fig:model}
\end{figure*}

\section{Related Work}
\paragraph{Sentence rewriting}
IUR can be considered to be similar to the sentence rewriting task \cite{Xu2021AttackingTC,Lin2021TowardsDP,Chen2018FastAS, cao2018retrieve}. Recent studies have been addressed the IUR task with various sophisticated methods. For example, \citeauthor{Pan-Restoration-EMNLP-19} \citeyear{Pan-Restoration-EMNLP-19} introduced a pick-then-combine model for IUR. The model picks up omitted tokens which are combined with incomplete utterances for restoration. \citeauthor{Liu-RUN-BERT-EMNLP-20} \citeyear{Liu-RUN-BERT-EMNLP-20} proposed a semantic segmentation method that segments tokens in an edit matrix then applied an edit operation to generate utterances. \citeauthor{Huang-SARG-AAAI-21} \citeyear{Huang-SARG-AAAI-21} presented a complicated model which uses a tagger for detecting kept, deleted, or changed tokens for restoration. We share the idea of using a tagger with \citeauthor{Huang-SARG-AAAI-21} \citeyear{Huang-SARG-AAAI-21} for IUR. However, we design a more simple but effective model which includes a picker (picking omitted tokens) and a generator for the restoration of incomplete utterances.




\paragraph{Text generation}
IUR can be formulated as text generation by using the seq2seq model \cite{Pan-Restoration-EMNLP-19,Huang-SARG-AAAI-21}. For the generation, several well-known pre-trained models have been applied \cite{Lewis2020BARTDS, Brown2020LanguageMA, Raffel2020ExploringTL} with promising results. We employ the T5 model \cite{Raffel2020ExploringTL} as the main component to rewrite utterances. To address the problem of missing important tokens in model's rewriting, we enhance T5 by introducing a Picker and two labeling methods (Section \ref{sec:model}).






\section{The Utterance Restoration Model}
\subsection{Problem Statement}
This work focuses on the incomplete utterance restoration of conversations. Let $H = \{h_1, h_2, ..., h_m\}$ be the history of the dialogue (context), $U = \{u_1, u_2, ..., u_n\}$ is the incomplete utterance that needs to be re-written. The task is to learn a mapping function $f(H, U| \Theta) = R$, where $R=\{r_1, r_2, ..., r_k\}$ is the re-written version of $U$. The learning of $\Theta$ is composed by only using utterance generation (the generator) or the combination of two tasks: important token identification (the picker) and utterance generation (the generator).


\subsection{The Proposed Model}\label{sec:model}
Our model is shown in Figure \ref{fig:model}. The Picker receives the context to identify omitted tokens. The Generator receives incomplete utterances for restoration. The model jointly learns to optimize the two tasks. Our model distinguishes in three significant differences compared to PAC \cite{Pan-Restoration-EMNLP-19} and SARG \cite{Huang-SARG-AAAI-21}. First, our model bases on a sing pre-trained model for both picker and generator while other models (i.e. PAC and SARG) use different architectures for the two steps. This makes two advantages for our model. (i) Our design can be easily adapted to create a new unified model for different tasks by using a single generative LM \cite{Paolini-TANL-ICLR-21}. (ii) Our model can work well in several scenarios: extraction vs. abstraction (data characteristics) and full vs. limited training data (Section \ref{sec:results}). Second, we design a joint training process to implicitly take into account the suggestion from the picker to the generator instead of using a two-step model as PAC which explicitly copes extracted tokens from the Pick for generation. Our joint training model can reduce the error accumulation compared to the two-step framework. Finally, we design a heuristic approach to build important tokens, which enable the model to work on a wider range of datasets and scenarios.



\subsubsection{Input representation}\label{subsec:input}
As shown in Figure \ref{fig:model}, we introduced three kinds of special tokens into the input text; $[X1]$, $[X2]$ and $\verb|<|\verb|\|s\verb|>|$. $[X1]$ and $[X2]$ are our newly defined special tokens and $\verb|<|\verb|\|s\verb|>|$ is the EOS token in the T5's vocabulary. We inserted $[X1]$ at the end of each utterance in the context, $[X2]$ after the incomplete utterance and $\verb|<|\verb|\|s\verb|>|$ at the end of whole input. $[X1]$ and $[X2]$ convey two pieces of useful information to the model; the signal indicating the switch of speakers and the cue to distinguish whether the utterance is from context or incomplete utterance.

The embedding of each token in the entire input sequence $S = \{w_1,w_2,...,w_l\}$ was obtained as $x_i = WE(w_i) + PE(w_i)$. Here, \textit{WE} is \textit{word embedding} initialized from a pretrained model by using a wordpiece vocabulary. \textit{PE} is \textit{relative position embedding} representing the position of each token in the sequence. These embeddings were fed into the $L$ stacked Encoder of T5; $E^l = EncoderBlock(E^{l-1})$ where $E^0 = \{x_1, … , x_l\}$. $E^L$ is the contextual representation of the whole input used by Picker and Generator in next sections. 

\subsubsection{The Picker}
It is possible to directly use T5 \cite{Raffel2019-aw} for IUR. However, we empower T5 with a Picker to implicitly take into account information from important tokens. The idea of selecting important tokens was derived from \citet{Pan-Restoration-EMNLP-19}, in which the authors suggested the use of important tokens contributes the performance of utterance restoration. We extend this idea by designing an end-to-end model which includes important token identification and generation, instead of using the two-step framework as \citet{Pan-Restoration-EMNLP-19}.

Given the context and the incomplete utterance, the Picker identifies tokens that are included in context utterances but omitted in the incomplete utterance. We call these tokens as important tokens. 
However, no important tokens are originally provided except for Restoration 200k in four datasets (please refer to Table \ref{tab:data}). Besides, the form of important tokens could change after restoration such as from plural to singular or nouns to verbs (Figure \ref{fig:example}). To overcome this issue, we introduce a label creation method that automatically identifies important tokens from the context for restoration.



\paragraph{Important token identification}\label{para:token}
Since building a set of important tokens is time-consuming and important tokens are usually not defined in practical cases, we introduce a heuristic strategy to automatically construct important tokens. In the following processing, stop words in the context, incomplete utterances, and gold references are removed in advance, assuming that stop words are the out of scope of important tokens. In addition, we applied lemmatization and stemming, the process of converting tokens to their base or root form, to alleviate the spelling variants.

First, we extracted tokens, called ``clue tokens", that exist in gold but not in incomplete utterances. If some tokens in context are semantically similar to some of the clue tokens, we can naturally presume that these tokens in the context are cited as important tokens for the rewriting. Therefore, we performed scoring by the distance $d_{ij}$ between the word representations of $i$-th token in context $h_i$ and $j$-th clue tokens $c_j$; $d_{ij} = \rm{cosine\_sim}(h_i, c_j)$ where \rm{cosine\_sim}() is the score of Cosine similarity. We used word representations of $h_i$ and $c_j$ from fastText \cite{bojanowski2017enriching} trained on Wikipedia as a simple setting of our model.

According to the distance $d_{ij}$, we introduce two types of labels for the Picker, \textit{soft}$_i$ as soft labels and \textit{hard}$_i$ as hard labels.
\begin{equation*}
    soft_i = \max_{j} d_{ij}
\end{equation*}

\begin{equation*}
    hard_i =
    \begin{cases}
        1   &   \max_{j} d_{ij} = 1\\
        0   &   otherwise
    \end{cases}
\end{equation*}

Here, the $\max$ operation was applied based on the assumption that at most one clue token corresponds to a token in the context. 

Intuitively, the soft label method takes into consideration the cases that could not be handled by lemmatization and stemming, such as paraphrasing by synonyms, and reflects them as the importance score in the range of 0 to 1. On the other hand, the hard label is either 0 or 1 where an important token is defined only when there is an exact matching between the context tokens and the clue tokens in the form after lemmatization and stemming. We provide the two methods to facilitate important token identification.

 \paragraph{Important token selection}
The Picker takes encoded embeddings $E^{L}=\{E^{L}_{1}, ..., E^{L}_{l}\}$ and predicts the scores of the soft label or hard label corresponding to each input token.
\begin{equation*}
    p(y_i|E^{L}_{i}) = softmax(\rm{FNN}(\textit{E}^{\textit{L}}_{\textit{i}}))
\end{equation*}
where $\rm{FNN()}$ is the vanilla feedforward neural network, which stands for projecting encoded embedding to the soft label or hard label space.
Then cross-entropy was adopted as the loss function.
\begin{equation*}
    \mathcal{L}_{picker} = -\sum_{i=1}^{l}q_i\log{p(y_i|E^{L}_{i})}
\end{equation*}
where $q_i$ is the picker's label for the $i$-th input token. To optimize loss function $\mathcal{L}_{picker}$ is equal to minimize the KL Divergence if the label is a soft label. In the hard labeling case, we assign three types of tags for tokens by following the BIO tag format as a sequence tagging problem.

\subsubsection{The Generator}

We explore the restoration task by using Text-to-Text Transfer Transformer (T5) \cite{Raffel2019-aw}. This is because T5 provides promising results for the text generation task. We initialized transformer modules from T5-base, which uses 12 layers, and fine-tuned it for our IUR task.

For restoration, encoder’s representation $E^L$ was fed into a $L$ stacked decoder with cross attention. $D^l_{i} = DecoderBlock(D^{l-1}_{i}, E^L)$ where $D^{0}_{i}=R_{<i}$, with $R_{<i} = \{<s>, r_1...,r_{i-1}\}$ and $<s>$ is the SOS token. The probability $p$ of a token $t$ at the time step $i$ was obtained by feeding the decoder’s output $D^L$ into the softmax layer.
\begin{equation*}
p(t\mid R_{<i}, H, U) = softmax(linear(D^{L}_{i}))\cdot v(t)
\end{equation*}

Here, $v(t)$ is a one-hot vector of a token $t$  with the dimension of vocabulary size. The objective is to minimize the negative likelihood of conditional probability between the predicted outputs from the model and the gold sequence $R=\{r_1,r_2,...,r_k\}$.
\begin{equation*}
\mathcal{L}_{generator} = - \sum_{i=1}^{k} \log {p(r_i \mid R_{<i}, H, U)}
\end{equation*}

\subsubsection{Joint learning}

JET aims to optimize the Picker and the Generator jointly as a setting of Multi-Task Learning. Different from PAC \cite{Pan-Restoration-EMNLP-19} that directly copies extracted tokens to generation, JET can implicitly utilize knowledge from the Picker, in which the learned patterns of the Picker to identify important tokens can be leveraged by the Generator. It can reduce error accumulation in the two-step framework as PAC. The final loss of the proposed model is defined as follows.
\begin{equation*}
    \mathcal{L} = \alpha \mathcal{L}_{picker} + \mathcal{L}_{generator}
\end{equation*}
where the hyperparameter $\alpha$ balances the influence of the task-specific weight. Our simple setting enables us to implement minimal experiments to evaluate how much important token extraction makes the contribution to generation.


\section{Settings and Evaluation Metrics}
\paragraph{Data}\label{sec:data}
We conducted all experiments on four well-known datasets of utterance rewriting in Table \ref{tab:data}.




\begin{table}[!h]
\centering
\setlength{\tabcolsep}{3.5pt}
\caption{\label{tab:data} Four conversational datasets. \textbf{ext} is extraction and \textbf{abs} is abstraction. CN: Chinese; EN: English.}
\begin{tabular}{lccccc}
\hline
Data  & train & dev & test & type & lang \\ \hline
Restoration 200k  & 194k & 5k & 5k & ext & CN \\
REWRITE  & 18k & 0 & 2k & ext & CN \\
TASK  & 2.2k & 0 & 2k & ext & EN \\
CANARD & 32k & 4k & 6k & abs & EN \\ \hline
\end{tabular}
\end{table}

Restoration 200k \cite{Pan-Restoration-EMNLP-19} and REWRITE \cite{Su-TPtr-ACL-19} include Chinese conversations. TASK \cite{Quan-GECOR-EMNLP-IJCNN-19} and CANARD \cite{Elgohary-CANARD-EMNLP-19} are in English, in which CANARD includes English questions from QuAC \cite{Choi-QuAC-EMNLP-18}. The datasets range from extraction to abstraction challenging UIR models.





\begin{table*}[!h]
\centering
\setlength{\tabcolsep}{6.5pt}
\caption{The comparison of JET and T5. \textbf{Bold numbers} show statistically significant improvements with $p \leq 0.05$. \underline{Underline} is comparable (applied to Tables \ref{tab:strong-baselines} and \ref{tab:rewrite-task}). The results come from the hard label method.}\label{tab:t5-joint}
\begin{tabular}{clccccccc} \hline
Data  & Method      & ROUGE-1 & ROUGE-2 & BLEU-1 & BLEU-2 & f1   & f2   & f3   \\ \hline

\multirow{2}{*}{\begin{tabular}[c]{@{}l@{}}Restoration \\200k\end{tabular}} 
    & T5-base & 92.7 & 86.1 & 91.4 & 88.9 & 61.3 & 51.2 & 44.8 \\
    & JET   & \textbf{93.1}    & \textbf{86.9}    & \textbf{92.0}   & \textbf{89.6}   & \textbf{63.0} & \textbf{53.3} & \textbf{47.1} \\ \hline
    
\multirow{2}{*}{\begin{tabular}[c]{@{}l@{}}REWRITE\end{tabular}} 
    & T5-base & 95.5 & 90.3 & 92.8 & 90.5 & 89.0 & 82.1 & 77.2 \\
    & JET   & \textbf{95.8} & \textbf{90.6} & \textbf{93.5} & \textbf{91.2} & \textbf{89.8} & \textbf{82.7} & \textbf{77.5} \\ \hline
    
\multirow{2}{*}{\begin{tabular}[c]{@{}l@{}}TASK\end{tabular}} 
    & T5-base & 95.8 & \underline{91.7} & 93.9 & 92.6 & \underline{76.2} & 71.2 & 68.1 \\
    & JET   & \textbf{96.1} & \underline{91.8} & \textbf{94.3} & \textbf{93.0} & \underline{76.3} & \textbf{72.1} & \textbf{69.6} \\ \hline
    
\multirow{2}{*}{\begin{tabular}[c]{@{}l@{}}CANARD\end{tabular}} 
    & T5-base   & 83.9  & 70.2    & 77.8  & 70.8   & 56.2  & 44.6  & 39.3 \\
    & JET  & \textbf{84.3}    & \textbf{71.1}    & \textbf{78.8}   & \textbf{72.0}   & \textbf{57.3} & \textbf{45.9} & \textbf{40.7} \\ \hline
    
\end{tabular}

\setlength{\tabcolsep}{5pt}
\caption{The comparison of JET and strong baselines; For Restoration 200k, results were derived from \citet{Liu-RUN-BERT-EMNLP-20} and \citet{Huang-SARG-AAAI-21}. For CANARD, we reproduced strong baselines which output promising results on Restoration 200k. The results of RUN-BERT on CANARD were derived from the code of \citet{Liu-RUN-BERT-EMNLP-20}. The results of JET come from hard labels.}\label{tab:strong-baselines}\vspace{0.2cm}
\begin{tabular}{clccccccc} \hline
Data                              & Method      & ROUGE-1 & ROUGE-2 & BLEU-1 & BLEU-2 & f1   & f2   & f3   \\ \hline
\multirow{9}{*}{\begin{tabular}[c]{@{}l@{}}Restoration \\200k\end{tabular}} 
    & Syntactic   & 89.3    & 80.6    & 84.1   & 81.2   & 47.9 & 38.8 & 32.5 \\
    & CopyNet     & 89.0    & 80.9    & 84.7   & 81.7   & 50.3 & 41.1 & 34.9 \\
    & T-Ptr       & 90.1    & 83.0    & 90.3   & 87.4   & 51.0 & 40.4 & 33.3 \\
    & PAC         & 91.6    & 82.8    & 89.9   & 86.3   & 63.7 & 49.7 & 40.4 \\
    & s2s-ft & 91.4    & 85.0    & 90.8   & 88.3   & 56.8 & 46.4 & 39.8 \\
    & RUN & 91.0 & 82.8 & 91.1 & 88.0 & 60.3 & 47.7 & 39.3 \\
RUN-BERT    & RUN-BERT & \underline{92.4} & 85.1 & \underline{92.3} & \textbf{89.6} & \textbf{68.6} & \textbf{56.0} & \textbf{47.7} \\
    & SARG        & 92.1    & \underline{86.0}    & \underline{92.2}   & \textbf{89.6}   & 62.4 & 52.5 & 46.3 \\
    & JET   & \textbf{93.1}    & \textbf{86.9}    & \underline{92.0}   & \textbf{89.6}   & 63.0 & 53.3 & \underline{47.1} \\ \hline
    
\multirow{4}{*}{\begin{tabular}[c]{@{}l@{}}CANARD\end{tabular}} 
    & s2s-ft & 83.1     & 69.0     & \underline{78.6}    & \underline{71.2}   & 55.1  & 43.2  & 37.9  \\
    & RUN-BERT & 80.6 & 62.7 & 70.2 & 61.2 & 44.2 & 30.5 & 24.9 \\
    & SARG  & 80.3    & 63.7    & 70.5   & 62.7   & 42.7 & 30.5 & 25.9 \\
    & JET   & \textbf{84.3}    & \textbf{71.1}    & \textbf{78.8}   & \textbf{72.0}   & \textbf{57.3} & \textbf{45.9} & \textbf{40.7} \\ \hline
    
    
\end{tabular}\vspace{-0.2cm}
\end{table*}

\paragraph{Settings}
For running JET, we used AdamW with $\beta_1=0.9$, $\beta_2=0.999$ and a weight decay of $0.01$ with a batch size of 12 and learning rate of $5e^{-5}$. We used 3 $\rm{FFN}$ layers (dimension as 768, 256, 64) with $\rm{ReLu}$ as the activation function. The final dimension is 1 for soft labeling and 3 for hard labeling. We set $\alpha =1$ for the loss function. We applied beam search with the beam size of 8. For picker's label creation, we used stop words from NLTK for English and from stopwordsiso\footnote{https://pypi.org/project/stopwordsiso/} for Chinese. For lemmatization and stemming, NLTK's WordNetLemmatizer and PorterStemmer were adopted for English, while lemmatization and stemming were skipped for Chinese. The pre-trained model was T5-base (English\footnote{https://huggingface.co/t5-base} and Chinese\footnote{https://huggingface.co/lemon234071/t5-base-Chinese}). In the full training data setting (Section \ref{sec:train-full}), the epoch size of 6 was used for Restoration200k and CANARD and 20 for REWRITE and TASK. In the limited training data setting (Section \ref{sec:train-lim}), the epoch size of 20 was used for all four datasets (Table \ref{tab:data}). All models were trained on a single Tesla P100 GPU. 




\paragraph{Evaluation metrics}
We followed prior work \cite{Pan-Restoration-EMNLP-19,Elgohary-CANARD-EMNLP-19,Liu-RUN-BERT-EMNLP-20,Huang-SARG-AAAI-21} to use three different metrics for evaluation, including ROUGE-scores, BLUE-scores, and f-scores.

\section{Results and Discussion}\label{sec:results}

\subsection{Full Training Data Setting}\label{sec:train-full}
We provide two scenarios of comparison with full training data: comparison with T5 and comparison with non-generative LM models.

\paragraph{Comparison with T5}
We first compare our model against a strong pre-trained T5 model used for the generator as the first scenario. This scenario ensures fair comparison among strong pre-trained models for text generation and also shows the contribution of the Picker. Results in Table \ref{tab:t5-joint} show that JET is consistently better than T5 across all metrics on all four datasets. This is because the picker can pick up important omitted tokens, which are beneficial for restoration. These results prove joint learning can implicitly supports to capturing the hidden relationship between the picker and generator. Also, the promising results show that our labeling method can work in both extraction and abstraction datasets. The results of T5 are also competitive. The reason is that T5 \cite{Raffel2019-aw} was trained with a huge amount of data by using the generative learning process, which mimics the text generation task. As the result, it is appropriate for the restoration.

For other strong pre-trained models for text generation, we also test our joint learning framework with ProphetNet \cite{Qi-ProPhetnet-EMNLP-20} but the results are not good to report. We leave the comparison with UniLM \cite{Dong-UniLM-NISP-19} and ERNIGEN \cite{Xiao-ERNIE-GEN-IJCAI-20} as a minor future task due to no pre-trained models for Chinese.



\paragraph{Comparison with non-generative LM models}
We next challenge JET to strong baselines which do not directly use generative pre-trained LMs, e.g. T5 for restoration. This is the second scenario that ensures the diversity of our evaluation. We leave the comparison of our model with BERT-like methods (e.g. SARG and RUN-BERT by using the T5 encoder) as a minor future task. For Restoration 200k and CANARD, we use the following baselines. \textbf{Syntactic} is the seq2seq model with attention \cite{Kumar-syntactic-COLING-16}. \textbf{CopyNet} is a LSTM-based seq2seq model with attention and the copy mechanism \cite{Huang-SARG-AAAI-21}. \textbf{T-Ptr} employs transformer layers for encoder-decoder for restoration \cite{Su-TPtr-ACL-19}. \textbf{PAC} is the two-stage model for utterance restoration \cite{Pan-Restoration-EMNLP-19}. \textbf{s2s-ft} leverages specific attention mask with several fine-tuning method \cite{bao2021s2s}. \textbf{RUN-BERT} is an IUR model by using semantic segmentation \cite{Liu-RUN-BERT-EMNLP-20}. \textbf{SARG} is a semi autoregressive model for multi-turn utterance restoration \cite{Huang-SARG-AAAI-21}.

Table \ref{tab:strong-baselines} shows that JET outputs promising results compared to strong baselines. For Restoration 200k, JET is competitive with RUN-BERT, the SOTA for this dataset. For CANARD, JET is consistently better than the baselines. The improvements come from the combination of the picker and generator. It is important to note that RUN-BERT and SARG are behind our model significantly on the abstractive scenario (CANARD). It supports our statement in Section \ref{sec:intro}, in which the current strong models for IUR is overspecific for extractive datasets and their generality is limited.


We next report the comparison on REWRITE and TASK in another table due to a small number of evaluation metrics. Following \citet{Liu-RUN-BERT-EMNLP-20}, we compare our model with RUN and two new methods: GECOR1 and GECOR2.
\begin{table}[!h]
\centering
\setlength{\tabcolsep}{3.1pt}
\caption{The comparison of JET and strong baselines on REWRITE and TASK; EM is exact match, B is BLEU, and R stands for ROUGE.}\label{tab:rewrite-task}
\begin{tabular}{clcccc} \hline
Data                              & Method      & EM & B4 & R2 & f1  \\ \hline
\multirow{4}{*}{\begin{tabular}[c]{@{}l@{}}REWRITE\end{tabular}} 
    & RUN & 53.8 & 79.4 & 85.1 & NA \\
    & T-ptr+BERT & 57.5 & 79.9 & 86.9 & NA \\
    & RUN-BERT & 66.4 & 86.2 & \underline{90.4} & NA \\
    & JET & \textbf{69.1} & \textbf{86.6} & \underline{90.6} & \textbf{89.8} \\ \hline
    
\multirow{4}{*}{\begin{tabular}[c]{@{}l@{}}TASK\end{tabular}} 
    & GECOR1 & 68.5 & 83.9 & NA & 66.1 \\
    & GECOR2 & 66.2 & 83.0 & NA & 66.2 \\ 
    & RUN & 69.2 & 85.6 & NA & 70.6 \\  
    & JET & \textbf{79.6} & \textbf{90.9} & \textbf{91.8} & \textbf{76.3} \\\hline
    
\end{tabular}
\end{table}

Results from Table \ref{tab:rewrite-task} are consistent with the results in Tables \ref{tab:t5-joint} and \ref{tab:strong-baselines}. It indicates that our model outperforms the baselines on both TASK and REWRITE. For REWRITE, the EM (exact match) score of our model is much better than the baselines. It shows that the model can correctly restore incomplete utterances. These results confirm that our model can work well in the two scenarios over all four datasets.

\paragraph{Important token ratio}

We observed how many important tokens are included in prediction on Restoration 200k. To do that, we defined two metrics, \textit{pickup ratio} and \textit{difference}. \textit{pickup ratio} indicates the ratio of predictions that contains important tokens on test datasets. \textit{difference} indicates the difference the character length between the prediction and the gold. Ideally, larger \textit{pickup ratio} with smaller \textit{difference} is desirable. 

\begin{table}[!h]
\centering
\setlength{\tabcolsep}{5pt}
\caption{The pickup ratio and difference of T5 and JET on Restoration 200k.}\label{tab:ratio}
\begin{tabular}{lcccc}\hline
        &  pickup ratio (\%) & difference \\ \hline
T5      & 29.0 & 1.28 \\
Defined & 29.9 & 1.30 \\ 
Soft    & 26.6 & 1.28 \\ 
Hard    & 30.4 & 1.21 \\ \hline
\end{tabular}
\end{table}

Table \ref{tab:ratio} shows JET with hard labeling achieves better results on both metrics compared to single T5. This supports our hypothesis that the Picker contributes the Generator for the IUR task.

\subsection{Limited Training Data Setting}\label{sec:train-lim}
We challenge our model in the limited training data setting. This simulates actual cases in which only a small number of training samples is available. We trained three strong methods: \textbf{SARG} \cite{Huang-SARG-AAAI-21}, T5, and JET on 10\% of training data by using sampling. We could not run RUN-BERT due to errors in the original code.


\begin{table}[!h]
\centering
\setlength{\tabcolsep}{4.5pt}
\caption{The comparison with limited training data.}\label{tab:limited-data}
\begin{tabular}{clcccc} \hline
Data  & Method      & R2 & B2 & f1   & f2   \\ \hline
\multirow{3}{*}{\begin{tabular}[c]{@{}l@{}}Restoration \\200k\end{tabular}} 
    & SARG      & 82.8 & 87.4 & 52.4 & \textbf{40.1}  \\
    & T5    & 84.5 & 87.2 & 53.8 & 37.2  \\
    & JET    & \textbf{84.6} & \textbf{88.0} & \textbf{56.5} & 39.2  \\
    \hline
    
\multirow{3}{*}{\begin{tabular}[c]{@{}l@{}}REWRITE\end{tabular}} 
    & SARG      & 57.9 & 50.0 & 0.00 & 0.00  \\
    & T5    & 77.3 & 73.7 & 71.0 & 61.3  \\
    & JET    & \textbf{77.7} & \textbf{74.9} & \textbf{72.1} & \textbf{62.4}  \\
    \hline
    
\multirow{3}{*}{\begin{tabular}[c]{@{}l@{}}TASK\end{tabular}} 
    & SARG      & 76.4 & 44.5 & 12.0 & 0.14  \\
    & T5    & 86.0 & 85.8 & 52.7 & 48.3  \\
    & JET    & \textbf{87.0} & \textbf{86.9} & \textbf{55.7} & \textbf{51.2}  \\
    \hline
    
\multirow{3}{*}{\begin{tabular}[c]{@{}l@{}}CANARD\end{tabular}} 
    & SARG      & 58.6 & 46.4 & 41.8 & 25.9  \\
    & T5    & 68.5 & 70.0 & 54.1 & 42.2  \\
    & JET    & \textbf{69.2} & \textbf{71.3} & \textbf{55.9} & \textbf{43.6}  \\
    \hline

\end{tabular}\vspace{-0.2cm}
\end{table}

As shown in Table \ref{tab:limited-data}, JET is consistently better than SARG with large margins. This is because JET is empowered by T5 which helps our model to work with a small number of training samples. This point is essential in actual cases. JET is also better than T5, showing the contribution of the Picker. SARG is good at ROUGE-scores and BLUE-scores but worse at f-scores, e.g. on REWRITE. The reason is that SARG uses the pointer generator network, that directly copies input sequences for generation, but it learns nothing.

\subsection{Soft Labels vs. Hard Labels}\label{sec:ablation}


    
    
    

We investigated the efficiency of our labeling method in Section \ref{para:token}. We run JET with soft and hard labeling methods. We also include the results of the JET on defined labels of Restoration 200k because this dataset originally provides labels of important tokens.

\begin{table}[!h]
\centering
\setlength{\tabcolsep}{4pt}
\caption{Soft vs. hard labeling methods. \textit{Defined} is the ground truths. T5 does not use the labeling methods.}\label{tab:soft-hard}
\begin{tabular}{clcccc} \hline
Data  & Method      & R2 & B2 & f1   & f2   \\ \hline

\multirow{4}{*}{\begin{tabular}[c]{@{}l@{}}Restoration \\200k\end{tabular}} 
    & T5      & 86.1 & 88.9 & 61.3 & 51.2  \\
    & Defined & 85.7 & 89.3 & 61.5 & 51.9  \\
    & Soft    & 86.2 & 87.9 & 57.4 & 47.7  \\
    & Hard    & \textbf{86.9} & \textbf{89.6} & \textbf{63.0} & \textbf{53.3}  \\ \hline
    
\multirow{2}{*}{\begin{tabular}[c]{@{}l@{}}CANARD\end{tabular}} 
    & T5      & 70.4 & 71.1 & 56.7 & 44.8  \\
    & Soft    & 70.8 & 71.4 & 57.1 & 45.5  \\
    & Hard    & \textbf{71.1} & \textbf{71.8} & \textbf{57.6} & \textbf{45.6}  \\ \hline
    
    
\end{tabular}
\end{table}
From Table \ref{tab:soft-hard} we can see the hard labeling method performs well on both datasets. Interestingly, the hard labeling method is even better than the one with defined labels on Restoration 200k. Although defined labels were manually created, Restoration 200k defines at most one important token in one sample even though some samples actually contain two or more omitted tokens. We found the hard label method detects 164k omitted tokens while the originally defined tokens are about 120k, and tokens detected by hard labeling cover 42\% of defined tokens. This suggests the hard label method extensively picks up important tokens even some important tokens are missing, and it can contribute to the enhancement of the JET.

For the soft labeling method, it contributes to the f-scores on CANARD (=abstractive) while it exacerbates accuracy on Restoration 200k (=extractive). This implies soft label does not function well in the distinction case between important and unimportant tokens is clear as in Restoration 200k. The soft labeling method would need more exploration on abtractive scenarios that require more synonymous paraphrasing or creative summarization.

\subsection{Human Evaluation}\label{sec:human-eval}
We report human evaluation with strong methods on CANRD because it is much more challenging than others. We asked three annotators who are well skilled in English and data annotation from the annotation team in our company. For the evaluation, we randomly selected 300 outputs from four models. Each annotator read each output and gave a score (1: bad; 2: acceptable; 3: good). Following \citet{kiyoumarsi2015evaluation} we adopted \textbf{Text flow} and \textbf{Understandability} as our criteria. \textbf{Text flow} shows how the restoration utterance is correct grammatically and easy to understand. \textbf{Understandability} shows how much the predictions are similar to reference semantically.
\begin{table}[!h]

\centering
\setlength{\tabcolsep}{3.5pt}
\caption{Human evalution on CANARD.}\label{tab:human-eval}
\begin{tabular}{l|cccc} \cline{2-5}
                  & SARG & s2s-ft & T5 & JET \\ \hline
Text flow         &  2.583    &2.887   &  2.925   & \textbf{2.933}     \\
Understand & 2.168     &  2.451  & 2.458     & \textbf{2.496}    \\ \hline
\end{tabular}
\end{table}

As shown in Table \ref{tab:human-eval}, JET obtains the highest scores on two criteria over other methods. It is consistent with the results of automatic evaluation in Tables \ref{tab:t5-joint} and \ref{tab:strong-baselines}. This is because our model utilizes strong pre-trained weights which provide the ability of text generation on unseen tokens, especially for abstractive data. The scores of JET also show the contribution of the Picker compared to the T5 for restoration.

\subsection{Output Observation}\label{sec:output-obs}
We observed the restoration outputs of different models in Figure \ref{fig:example}. There exist 9 omitted tokens between the incomplete utterance and the reference. The SARG and s2s-ft can restore only 2 important tokens. T5 can restore 8 the important tokens out of 9 but generates unnecessary words. Our proposed model also can restore 8 important tokens and have the same semantic meaning as the gold utterance. This suggests our model learns to use only the tokens picked up by Picker as additional tokens for rewriting.

\begin{table}[!h]
\centering
\setlength{\tabcolsep}{5pt}
\caption{The average BLEU score on CANARD.}\label{tab:blue-length}
\begin{tabular}{l|cccc} \hline
 Length        & < 100 & 100 $\leq$ 200 & 200 <  \\ \hline
 SARG         &  55.53    &45.46  & 38.96      \\
 s2s-ft         &  63.89    &54.94   & 48.39       \\
 T5        & \textbf{65.03}    &55.69   &  51.25     \\
 JET	     &64.94     &\textbf{ 56.56 } & \textbf{52.84}   \\ \hline
\end{tabular}
\end{table}
We also examined the ability of strong methods with different input lengths on CANARD. Results in Table \ref{tab:blue-length} show that our model can deal with longer input sequences. Compared to SARG and seq2seq, the performance of our model is much better. This is because the implicit suggestion from the Picker combined with the ability to deal with long sequences of T5 increase the score.

\section{Conclusion}
This paper introduces a simple but effective model for incomplete utterance restoration. The model is designed based on the nature of conversational utterances, where important omitted tokens should be included in restored utterances. To do that, we introduce a picker with two labeling methods for supporting a generator for restoration. We found that the picker contributes to improve the generality of the model on four benchmark datasets. The model works well in English and Chinese, from extractive to abstractive scenario in both full and limited training data settings. The future work will investigate the behavior of the model in other domains and the potential application of JET, e.g. combining utterance extraction and utterance restoration for information extraction from dialogue.
\section*{Acknowledgement}
We would like to thank Yun-Nung Chen and anonymous ACL ARR reviewers who gave constructive comments for our paper.

\bibliography{conversation,conversation_ryan}
\bibliographystyle{acl_natbib}




\end{document}